\documentclass[journal]{IEEEtran}
\usepackage{amsmath,graphicx}
\usepackage{amssymb}
\usepackage[hang,center]{subfigure}
\usepackage{tabularx}
\usepackage{array}
\usepackage{bm}
\usepackage{epsfig}
\usepackage{algorithmic}
\usepackage{balance}
\usepackage[lined,linesnumbered,ruled]{algorithm2e}
\usepackage{url}
\usepackage{xcolor}
\usepackage[footnotesize]{caption}
\usepackage{multirow}
\usepackage{rotating}
\usepackage{lscape}
\graphicspath{{../../Figs/}}
\newcommand{\minitab}[2][l]{\begin{tabular}{#1}#2\end{tabular}}
\ifCLASSINFOpdf
\else
\fi
\begin{document}
%
\title{Robustness Analysis of Pedestrian Detectors \\ for Surveillance}

\author{Yuming~Fang,~\IEEEmembership{Senior Memmber,~IEEE},
        Guanqun Ding,
        Yuan~Yuan,
        Weisi~Lin,~\IEEEmembership{Fellow,~IEEE}, \\
        and Haiwen Liu,~\IEEEmembership{Senior Memmber,~IEEE}
\thanks{This work was supported in part by the National Natural Science Foundation of China under Grant 61571212, the Fok Ying-Tong Education Foundation of China under Grant 161061, and the Natural Science Foundation of Jiangxi under Grant 20071BBE50068 and 20171BCB23048.}
\thanks{Yuming Fang and Guanqun Ding are with the School of Information Technology, Jiangxi University of Finance and Economics, Nanchang, Jiangxi, China, E-mail: fa0001ng@e.ntu.edu.sg.}
\thanks{Yuan Yuan and Weisi Lin are with the School of Computer Engineering, Nanyang Technological University, Singapore, 639798. Email: \{yyuan004,wslin\}@ntu.edu.sg.}
\thanks{Haiwen Liu is with the School of Electronic and Information Engineering, Xi$'$an Jiaotong University, Xi$'$an 710049, China. Email: liuhaiwen@gmail.com}}

%

\maketitle

\begin{abstract}
To obtain effective pedestrian detection results in surveillance video, there have been many methods proposed to handle the problems from severe occlusion, pose variation, clutter background, \emph{etc}. Besides detection accuracy, a robust surveillance video system should be stable to video quality degradation by network transmission, environment variation, \emph{etc}. In this study, we conduct the research on the robustness of pedestrian detection algorithms to video quality degradation. The main contribution of this work includes the following three aspects. First, a large-scale Distorted Surveillance Video Data Set (\emph{DSurVD}) is constructed from high-quality video sequences and their corresponding distorted versions. Second, we design a method to evaluate detection stability and a robustness measure called \emph{Robustness Quadrangle}, which can be adopted to visualize detection accuracy of pedestrian detection algorithms on high-quality video sequences and stability with video quality degradation. Third, the robustness of seven existing pedestrian detection algorithms is evaluated by the built \emph{DSurVD}. Experimental results show that the robustness can be further improved for existing pedestrian detection algorithms. Additionally, we provide much in-depth discussion on how different distortion types influence the performance of pedestrian detection algorithms, which is important to design effective pedestrian detection algorithms for surveillance.


\end{abstract}


%
\IEEEpeerreviewmaketitle

\section{Introduction}
\label{sec:intro}
Pedestrian detection plays an important role in auto-analysing of surveillance video. It is the prerequisite of various tasks of surveillance video processing including pedestrian tracking, crowd analysis, event recognition, anomaly detection, \emph{etc}. During the last decade, significant progress has been achieved on existing published data sets including Caviar \cite{Caviar}, INRIA \cite{dalal2005histograms}, Caltech \cite{dollar2012pedestrian}, PETS09 \cite{ferryman2009PETS}, TUD-Stadtmitte \cite{andriluka2010}, \emph{etc}. \cite{Sszhang2016}. These data sets challenge the pedestrian detection algorithms by introducing different levels of occlusion, dynamic shape variation, different aspect ratios, \emph{etc} \cite{Ouyang2017}. By addressing these content-related challenges, various pedestrian detection algorithms \cite{benenson2013seeking,ShenJF2017,dollar2014ACF,TianYL2015,wang2009hoglbp, sakrapee2016,wojek2009multi,wu2011real,CaoJL2016,zhang2014informed,PengP2015} have been designed to obtain higher detection accuracy. It is reported \cite{Benenson2014Eccvw} that the log-average miss-rate has decreased from around $70\%$ \cite{dalal2005histograms} to around $35\%$ \cite{zhang2014informed} on Caltech data set \cite{dollar2012pedestrian}.


However, in surveillance systems, the quality of surveillance video may change from time to time due to varies factors such as, bandwidth limitation, illumination variation, sensor variety of different cameras, \emph{etc}. \cite{LiLD2016, GUKe2016}. When video quality decreases, targets may not be distinguishable any more in the distorted video, and this results in wrong detection. Thus, the effect of video quality variation on pedestrian detection should be investigated. There have been several studies focusing on assessing the quality of distorted image/video for face and event detection \cite{korshunov2011video,kafetzakis2013impact}. It has been demonstrated that detectors always favor high quality image/video to obtain promising detection accuracy. On the other hand, a robust system requires detection algorithms which perform robustly and accurately in different quality conditions as well. There are some studies investigating into the benchmark of pedestrian detection \cite{dollar2012pedestrian, Dollar2009CVPR}. In these studies, the authors build a large-scale database to study the statistics of the size, position and occlusion patterns of pedestrians in urban scenes. A new per-frame evaluation method is designed to measure the performance of different pedestrian detection algorithms. However, they do not consider the performance robustness of different pedestrian detection methods for quality-degradation video sequences. Currently, there is no systematic study focusing on the robustness of pedestrian detectors regarding to surveillance video quality variation, which motivates us to build a Distorted Surveillance Video Data Set (\emph{DSurVD}) and study the robustness of pedestrian detectors to video quality degradation in surveillance systems. Our initial work has been reported in \cite{YuanY2015ICIP}.

\begin{figure*}[htb]
  \centering
  \includegraphics[width=0.96\textwidth]{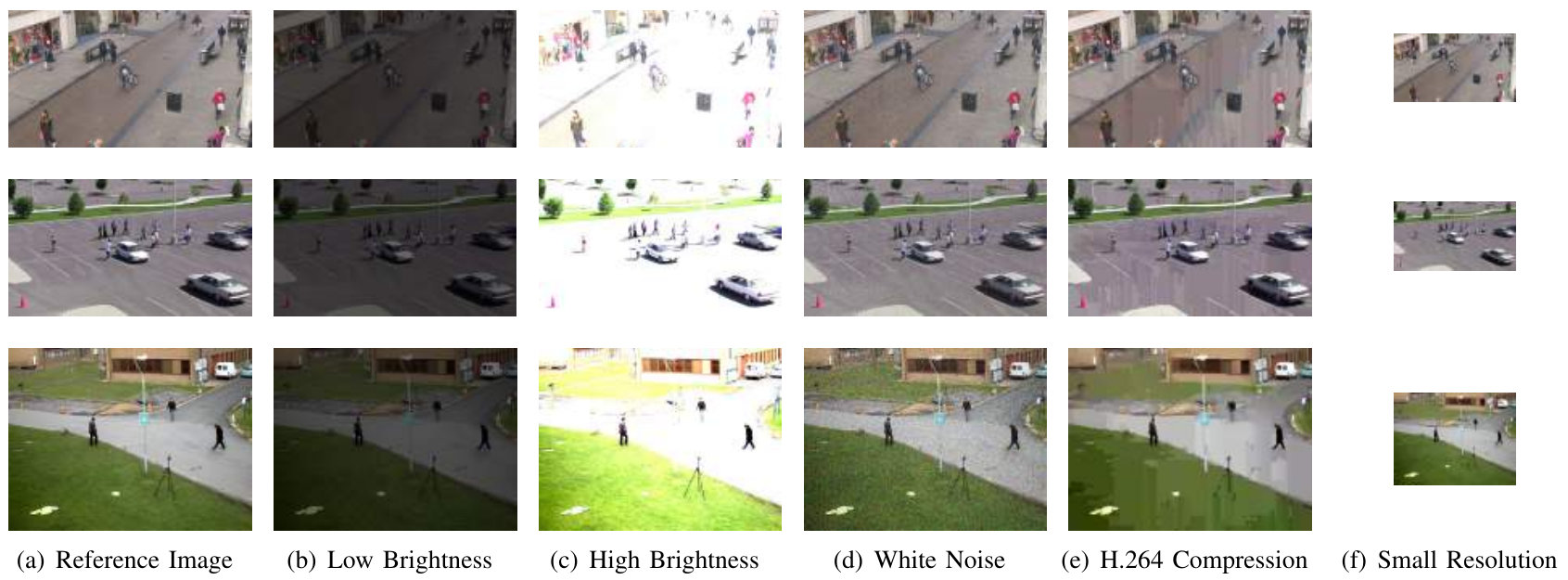}
  \caption{Sample images in \emph{DSurVD}, with four types of distortion introduced. Column (a) are the reference image frames with high quality; columns (b) and (c) are image frames with brightness variations; column (d) are image frames with additive white noise; column (e) are image frames with quality degradation after H.264 compression; column (f) are image frames with lower resolution. }\label{fig:ImgSamples}
\end{figure*}

Generally, distortion in surveillance video may be caused by bandwidth limitation, noise in video acquisition, brightness variation due to camera variety, illumination change, etc. In this study, we consider the following four distortion types in the proposed \emph{DSurVD}: compression distortion, resolution reduction, white noise and brightness changes. Regarding bandwidth limitation, distortion of video is mainly from compression distortion or resolution reduction. We also introduce different levels of white noise to the high-quality reference videos to obtain the noisy videos. Moreover, we adjust the brightness of the video to obtain the corresponding distorted versions with both high brightness and low brightness. Three common surveillance scenes are considered in the proposed \emph{DSurVD}, including campus, town centre and car park. Fig. \ref{fig:ImgSamples} illustrates some sample video frames in \emph{DSurVD}.

Furthermore, we evaluate 7 existing pedestrian detectors which are published in studies \cite{wojek2009multi,wang2009hoglbp,wu2011real,dollar2014ACF,dalal2005histograms} on the proposed \emph{DSurVD}. To study the robustness of detectors, both the detection accuracy $A_\mathrm{ref}$ on high-quality reference videos and performance stability $\mathbf{S}$ on distorted videos are measured. With $A_\mathrm{ref}$ and $\mathbf{S}$, we define the robustness quadrangles (seen in Fig. \ref{fig:RobustnessQuad}) to visualize the robustness of different detectors. Based on the proposed robustness quadrangle, we know the advantages and disadvantages of detectors regarding to detection accuracy and stability with certain distortion type. Based on the in-depth analysis of the stability of existing pedestrian detectors with different distortion types, we facilitate some possibilities to improve the detection stability of pedestrian detectors regarding to video quality degradation.

The rest paper is organized as follows. We introduce the proposed \emph{DSurVD} and the statistics of the distorted video sequences in Section \ref{sec:DSurVD}. Section \ref{sec:AccySta} provides the definition of detection robustness; and a detection stability measurement is proposed (as Section \ref{subsec:Stability}). In Section \ref{sec:EvalResults}, the evaluation of pedestrian detectors on \emph{DSurVD} is reported; the in-depth analysis is given in this section as well. Finally, we summarize the robustness study of pedestrian detection in surveillance video and discuss the possibilities in the future research in Section \ref{sec:Discussion}.


\section{Distorted Surveillance Video Data Set}
\label{sec:DSurVD}
In surveillance systems, due to bandwidth and storage limitation, video quality may vary after compression. Moreover, different environments, camera variety and unpredictable noise during video acquisition may influence the video quality as well. In this study, in order to study the performance of pedestrian detectors in surveillance video with quality variation, the first Distorted Surveillance Video Data Set (\emph{DSurVD}) containing video sequences with distortion versions from different quality levels is constructed.

With H.264/AVC \cite{wiegand2003overview}, one of the most widely used video coding standards for surveillance video, it is easy to adjust the \emph{quantization parameter} (QP), \emph{resolution} and frame rate to meet the limitation of the bandwidth. In general, if the QP and resolution are fixed, changing the frame rate does not significantly affect the quality of individual frames in a video sequence. In addition, since most pedestrian detectors process each video frame separately, without changing QP and resolution, frame rate does not affect detection accuracy. In the proposed \emph{DSurVD}, video sequences with different QPs and resolutions are created as distorted versions. Moreover, additive noise during video acquisition and brightness variation caused by illumination change or overexposing are two important distortion sources. Hence, two more distortion types of \emph{white noise} and \emph{brightness variation} are included with \emph{DSurVD}.

\subsection{Reference Video Sequences}
\label{subsec:ReferenceVideo}
In \emph{DSurVD}, the distorted video sequences are created based on five high quality surveillance video sequences including scenarios of campus (two sequences in PETS09 \cite{ferryman2009PETS}), town centre (TownCentre sequence \cite{benfold2011stable}) and car park (ParkingLot1 and ParkingLot2 sequences \cite{shu2013improving}). The ground truth are manually labeled bounding boxes of pedestrians.

These five video sequences are typical surveillance video data which have been widely used in recent pedestrian detection and tracking studies \cite{milan2013detection,yan2012multi,benfold2011stable,tang2013learning,yuan2014tracking}. Furthermore, these sequences are with relatively high resolution and constant pedestrian size, and are captured with fixed cameras. Captured with fixed cameras guarantees relatively stable quality of all the frames in each video sequence. The reason why we prefer constant pedestrian size is as follows. As we reduce the resolution of reference videos, the lose of high frequency information of pedestrians caused by pedestrian size reduction is the main factor that affect the performance of detection algorithms. Thus, in order to study the relationship between the pedestrian size and detection accuracy, it is better to have a constant pedestrian size in the reference video. We use the variation coefficient of the pedestrian height ($H_{vc}$) to represent the variation of pedestrian size:
\begin{equation}\label{eq:HgtCV}
  H_{vc} = \frac{\sigma_{h}}{\mu_{h}},
\end{equation}
where $\sigma_{h}$ and $\mu_{h}$ denote the standard deviation and the mean of pedestrian height (in pixels) respectively, which are obtained from the ground truth of each data set. From Eq. \ref{eq:HgtCV}, we can see that a smaller value of $H_{vc}$ indicates more constant pedestrian size. Table \ref{tab:SeqStat} provides the $H_{vc}$ and the resolution of some popular pedestrian detection/tracking video data sets. It shows that the five reference sequences used in \emph{DSurVD} are with highly constant pedestrian size and relatively high resolution.

\begin{table*}[!htbp] \centering \small
\caption{Mean and coefficient of variation $H_{vc}$ of pedestrian heights (in pixel); frame resolution of several existing pedestrian detection/tracking video data sets; the last row indicates whether the video sequences are captured by a fixed camera or not. (See Section \ref{subsec:ReferenceVideo} for details)}
  \label{tab:SeqStat} 
\begin{tabular}{|c|c|c|c|c|c|c|c|}
  \hline
    & PETS09\cite{ferryman2009PETS} & ParkingLot\cite{shu2013improving} & TownCentre\cite{benfold2011stable} & TUD-Cam.\cite{andriluka2008people} & Caviar\cite{Caviar} & Caltech\cite{dollar2012pedestrian} & ETH\cite{ess2008mobile} \\
  \hline
   $\mu_{h}$ & 83.8 & 171.6 & 203.6 & 207.5 & 64.1 & 51.4 & 254.2 \\
  \hline
   $H_{vc}$ & 0.23 & 0.09 & 0.31 & 0.27 & 0.43 & 0.65 & 0.66 \\
   \hline
   Res. & $768\times576$ & $1920\times1080$ & $1920\times1080$ & $640\times480$ & $384\times288$ & $640\times480$ & $640\times480$ \\
  \hline
   FixCam & $\surd$ & $\surd$ & $\surd$ & $\surd$ & $\times$ & $\times$ & $\times$ \\
  \hline
\end{tabular}
\end{table*}

\subsection{Distorted Video Sequences}
\label{subsec:DistortedVideo}
For each reference sequence, we create $52$ distortion versions (as explained next) based on the aforementioned four distortion types. Hence, including the reference sequences, there are $53\times5=265$ video sequences in total in the \emph{DSurVD}. Below, we analyze the statistics of the \emph{DSurVD} in detail.



\begin{figure*}[!htbp]
  \centering
  \includegraphics[width=0.96\textwidth]{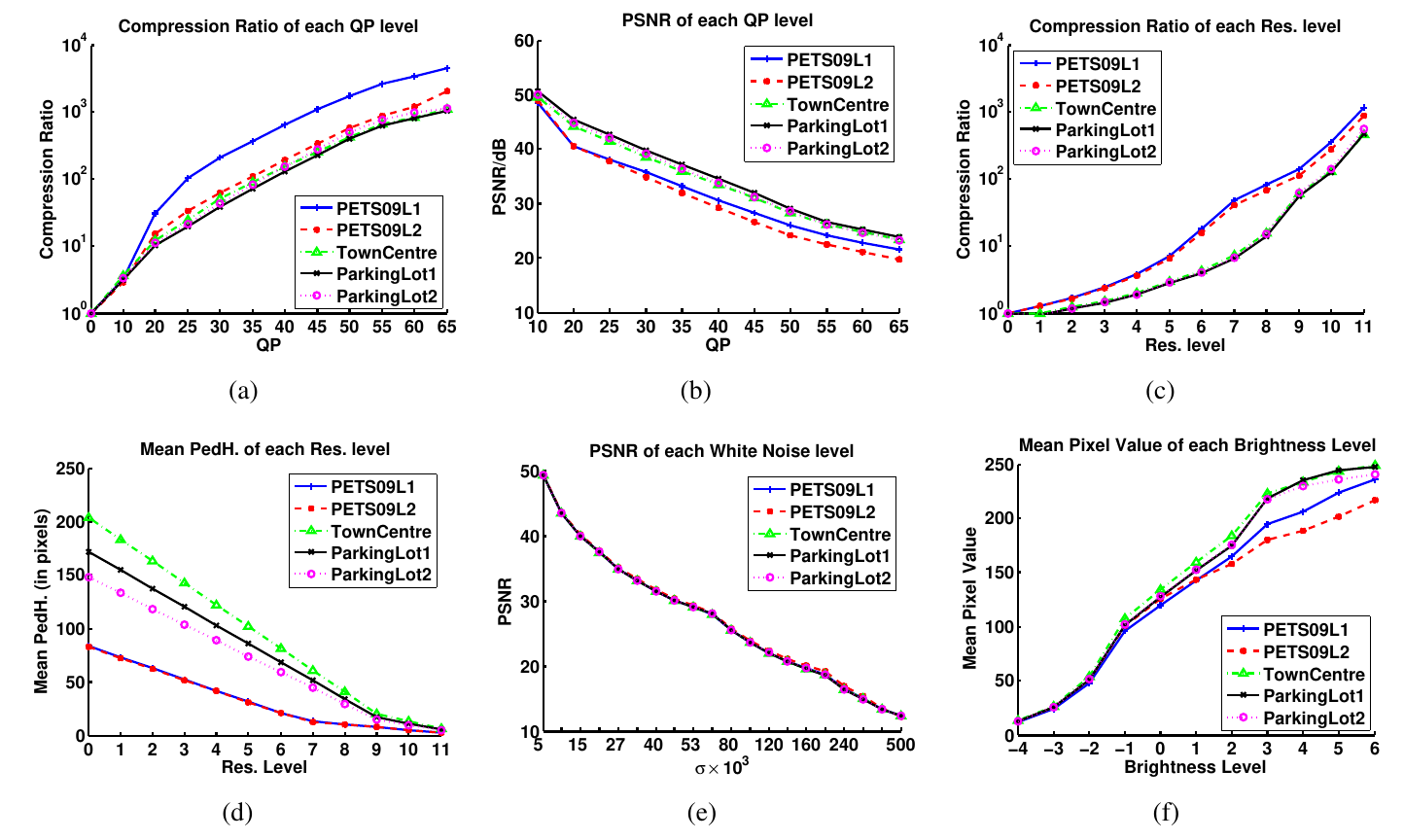}
  \caption{(a) and (b) are the compression ratio and PSNR statistics of the distorted videos by adjusting the QP; (c) and (d) are the compression ratio and mean pedestrian height of the distorted videos by adjust the resolution, higher resolution level indicates smaller resolution of the video and $0$ refers to the reference videos; (e) is the PSNR statistics of the distorted videos by adding zero mean Gaussian noise, where x-axis is the $\sigma$ of the Gaussian kernel; (f) is the mean pixel value of distorted video with brightness variation, ``0'' indicates the reference video.(See Section \ref{subsec:DistortedVideo} for details)}\label{fig:PSNR_QP_WN}
\end{figure*}

\textbf{Quantization Parameter:} QP is one of the most important parameters in H.264 codec to encode video stream with different bit rates. The quality of the video is degraded by increasing the value of QP. In H.264, the Quantization Parameter (QP) determines the quantization step of the transformed coefficients with Discrete Cosine Transform. Larger QP refers to the bigger step and results in poorer video quality while lower QP refers to the smaller step and results in better quality. QP cannot directly refer to the bitrate since the bitrate is content biased. However, in general, each unit increase of QP lengthens the step size by 12\% and reduces the bitrate by roughly 12\% in H.264. Detailed information can be referred to the study \cite{Schwarz2007}. In \emph{DSurVD}, we encode each reference video sequence with $11$ quality levels by varying QP from $10$ to $65$. These distorted sequences are named as SqsQP. The codec we used is ffmpeg \cite{bellard2014ffmpeg}.

The peak signal noise ratio (PSNR) of each distorted sequence is computed and shown in Fig. \ref{fig:PSNR_QP_WN}. Within the $11$ distortion levels, PSNR drops from $\sim50$dB to $\sim23$dB. PSNR of PETS09L1 \& 2 is a little lower than that of TownCentre and ParkingLot1 \& 2, due to the distortion in the large grass regions which are with more complex texture in PETS09L1 \& 2.

The compression ratio of each distorted sequence is computed and shown in Fig. \ref{fig:PSNR_QP_WN}. The compression ratio varies from $1$ to around $10^3$ between the reference sequences and the distorted sequences. The compression ratio of PETS09L1 is higher than that of PETS09L2 with the same QP even when they have the same background scene. The reason is the low density of pedestrians in PETS09L1. Based on the statistics of manually labelled ground truth, the average number of pedestrians per frame in PETS09L1 is $5.8$ which is much lower than $23.6$ in PETS09L2. Lower pedestrian density may result in less motion in the video, and thus less bite rate is required for inter coding between consecutive frames with H.264 codec.

\textbf{Resolution:} Reducing the resolution is an alternative way to meet low bandwidth limitation in H.264 codec. In \emph{DSurVD}, we code $11$ video sequences with low resolution for each reference video sequence. These sequences are named as SqsRes.

To video sequences PETS09L1 \& 2, the resolution is reduced from $768\times576$ (reference video) to $24\times18$. For video sequences TownCentre and ParkingLot1 \& 2, the resolution is reduced from $1920\times1080$ (reference video) to $64\times36$. The compression ratio of each distorted sequence is computed and shown in Fig. \ref{fig:PSNR_QP_WN}. ``$0$'' in the Res. Level axis indicates the reference video sequences and ``$11$'' indicates the video sequences with the lowest resolution ($24\times18$ for PETS09L1 \& 2, and $64\times36$ for TownCentre and ParkingLot1 \& 2).

Pedestrian size change is the direct effect by reducing the resolution of video sequences. As mentioned in \cite{dollar2012pedestrian}, pedestrian height cannot be neglected for pedestrian detection accuracy. We plot the relation between the average pedestrian height and the resolution level of each sequence in Fig. \ref{fig:PSNR_QP_WN}. The average pedestrian height (in pixels) varies from around $200$ to about $10$ in the proposed \emph{DSurVD}. From level ``$0$'' to level ``7'', the down sampling step of each video is kept the same, and it can be seen from Fig. \ref{fig:PSNR_QP_WN} that the slope of the each line keeps the same before level ``7'' . In order to study more detail of pedestrian detectors with low resolution (or small pedestrian size), we decrease the down sampling step after level ``7'' to get more low resolution videos, and we can see that the slope of each line become smaller after level ``7''.

\textbf{White Noise:} Apart from the aforementioned distortion types introduced by compression, white noise is another common type of distortion during image/video acquisition \cite{ponomarenko2009tid2008}. We use the zero-mean Gaussian noise to model the additive white noise. In total, $20$ levels of Gaussian noise are added to the reference video sequences where the Gaussian Kernel $\sigma$ varies from $0.005$ to $0.5$, and the PSNR varies from $\sim50$dB to $\sim25$dB, respectively. These sequences are named as SqsWN. The PSNR of each noisy sequence is computed and shown in Fig. \ref{fig:PSNR_QP_WN}. The PSNR is highly correlated to $\sigma$ and there is almost no difference of PSNR between different reference video sequences.

\textbf{Brightness Variation} The brightness variation of video frames in surveillance video sequences can be caused by both illumination change and different exposure sensitivity of the camera. We model $10$ levels of brightness for video sequences in \emph{DSurVD}. These sequences are named as SqsBV. In total, the distortion versions with $4$ low and $6$ high brightness levels are created. The mean pixel value of each brightness level are shown in Fig. \ref{fig:PSNR_QP_WN}.


\section{Robustness Analysis: Accuracy and Stability}
\label{sec:AccySta}
The IEEE Standard Glossary of Software Engineering Terminology \cite{radatz1990ieee} gives the definition of robustness as follows:
\\
\\
Robustness: \emph{The degree to which a system or component can function correctly in the presence of invalid inputs or stressful environmental conditions.}
\\
\\
Based on this definition, the robustness of the pedestrian detector can be measured from the following two aspects. On one hand, what is the accurate rate of the pedestrian detector in surveillance videos? On the other hand, what is the performance stability of the pedestrian detector with video quality degradation? In this section, we denote these two aspects as accuracy and stability, respectively, and propose an approach to measure them.

\subsection{Accuracy Measurement}
\label{subsec:Accuracy}
Given a detection bounding box $bb_{dt}$ and a ground truth bounding box $bb_{gt}$, we employ the matching criterion used in the previous pedestrian detection benchmark study \cite{dollar2012pedestrian}. With the overlap between $bb_{dt}$ and $bb_{gt}$ exceeding $50$ percent, we consider them as a correct match,
\begin{equation}\label{eq:overlap}
    \frac{area(bb_{dt} \bigcap bb_{gt})}{area(bb_{dt} \bigcup bb_{gt})}>0.5.
\end{equation}
Each $bb_{gt}$ can match to at most one $bb_{dt}$. If a detection bounding box matches multiple ground truth bounding boxes, the match with the highest overlap is used.
Unmatched $bb_{dt}$ and $bb_{gt}$ are considered as false positive samples and false negative samples, respectively.

We plot miss rate against false positives per image (FPPI) to visualize the performance of pedestrian detector (e.g., Fig. \ref{fig:AccuracyPlot}). Similar to \cite{dollar2012pedestrian}, the \emph{log-average miss rate} ($\mathcal{MR}$) is computed by averaging miss rate at nine FPPI rates evenly spaced in log-space in the range $10^{-2}$ to $10^{0}$, to quantify the performance of the detector. We define the \emph{Accuracy} by subtracting $\mathcal{MR}$ by $1$ to make sure it is positive related to performance:
\begin{equation}\label{eq:Accuracy}
    A = 1 - \mathcal{MR}.
\end{equation}
The value of $A$ ranges from $[0,1]$ and larger value indicates better performance of pedestrian detector. 

\subsection{Stability Measurement}
\label{subsec:Stability}
\emph{Accuracy} is an important measurement index for pedestrian detectors. In traditional pedestrian detection data sets \cite{everingham2010pascal}, different \emph{Accuracy} metrics have been proposed to measure the performance of pedestrian detectors. However, \emph{Stability} is another unneglectable measurement index for pedestrian detectors. In the case where two pedestrian detectors have similar \emph{Accuracy} on good quality-video sequences, the \emph{Stability} measurement provides another important dimension to evaluate the performance. Here, we propose a \emph{Stability} ($\mathbf{S}$) measurement method by analyzing the performance of pedestrian detectors in surveillance videos with quality degradation. To the best of our knowledge, this is the first study to provide the dedicated analysis of stability of pedestrian detectors for visual surveillance with video quality variation.

\begin{figure}
  \centering
  \includegraphics[width=0.98\columnwidth]{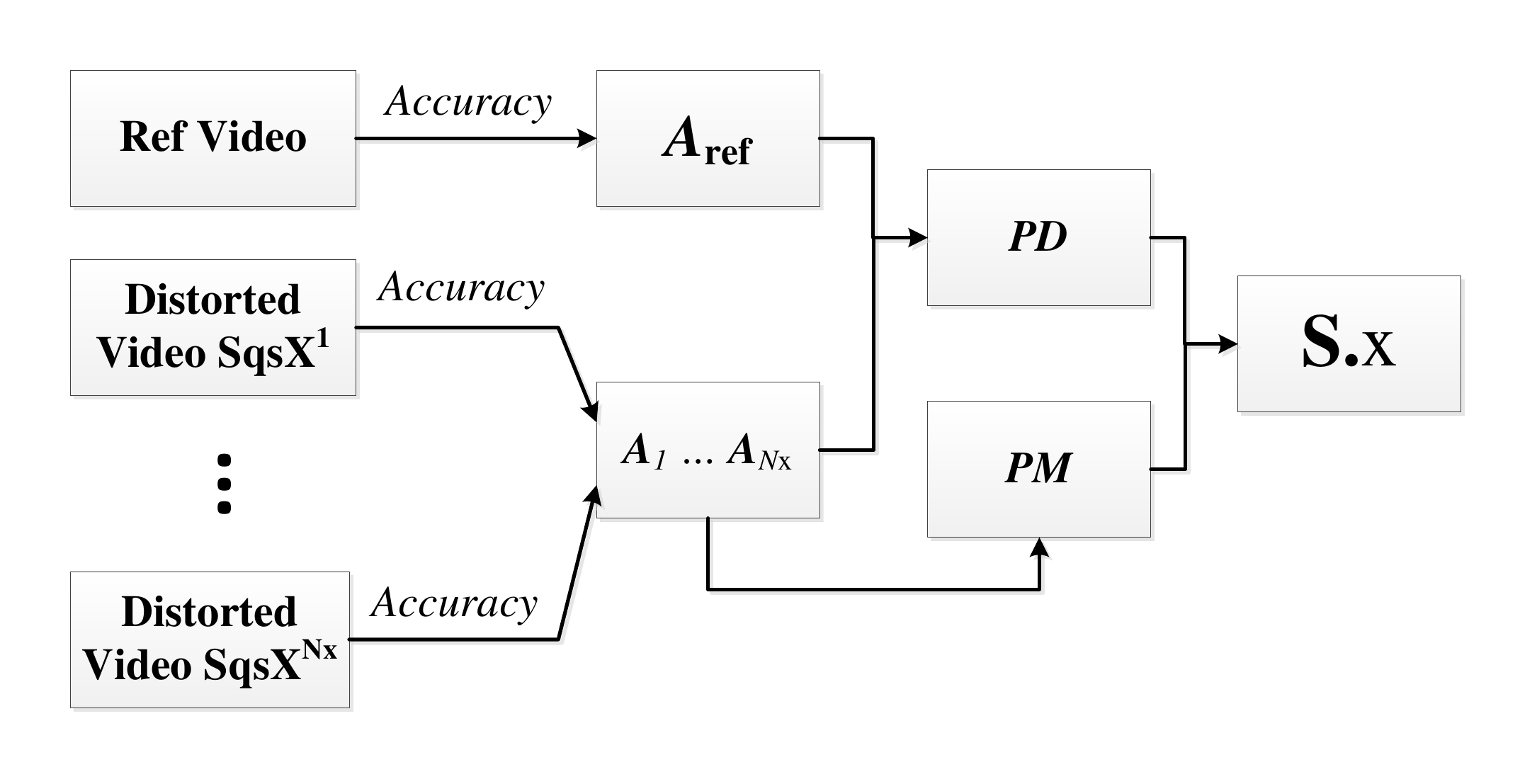}\\
  \caption{\emph{Stability} evaluation based on the variability of \emph{Accuracy} from the reference video to the most distorted video.
  $\mathrm{SqsX^{1}}$ to $\mathrm{SqsX^{N}}$ indicate the distorted video sequences with distortion type $\mathrm{x}$.}\label{fig:Accuracy2Stability}
\end{figure}

With the four aforementioned common distortion types, we define \emph{Stability} as a four dimensional vector,
\begin{equation}\label{eq:Stability}
\mathbf{S} = [\mathbf{S}.\mathrm{qp}, \mathbf{S}.\mathrm{res}, \mathbf{S}.\mathrm{wn}, \mathbf{S}.\mathrm{bv}],
\end{equation}
where $\mathbf{S}.\mathrm{qp}$, $\mathbf{S}.\mathrm{res}$, $\mathbf{S}.\mathrm{wn}$ and $\mathbf{S}.\mathrm{bv}$ denote the \emph{Stability} of pedestrian detectors with QP variation, resolution variation, additive white noise and brightness variation, respectively.

To quantify the \emph{Stability}, two criteria are incorporated in the study as follows:
\\
\\
\textbf{Rate of Accuracy Degradation:} The \emph{Accuracy} degradation rate of a robust detector should be slow when input video quality decreases.
\\
\\
\textbf{Monotonicity:} A robust detector would show a monotonically degradation in \emph{Accuracy} when input video quality decreases.
\\
\\
It is easy to understand the slow accuracy degradation rate criterion in degradation study. The motivation of the monotonicity criterion is that, with quality degradation, detectors whose detection accuracy oscillates are much less predictable than detectors with monotonically accuracy degradation. In other words, when increasing the video quality gradually, we prefer monotonically increasing of the detection accuracy rather than oscillating of the detection accuracy.

Given a reference video sequence and a particular distortion type $\mathrm{x}$ (e.g., $\mathrm{x}$ can be $\mathrm{qp}$, $\mathrm{res}$, $\mathrm{wn}$ and $\mathrm{bv}$), we first compute the detection \emph{Accuracy} values with the reference video sequence $A_\mathrm{ref}$ and all the distorted video sequences $\{A_i: i = 1,\ldots,N_\mathrm{x}\}$, where $N_\mathrm{x}$ is the number of distorted sequences with distortion type $\mathrm{x}$.

For the $i^{th}$ distorted sequence, we formulate the penalty of accuracy degradation $PD_i$:

\begin{equation}\label{eq:DegradationPenalty}
    PD_i = \min\{1,(\frac{A_i-A_\mathrm{ref}}{A_\mathrm{ref}})^2\},
\end{equation}
where $A_{i}$, and $A_\mathrm{ref}$ are the detection accuracy on the $i^{th}$ distorted video sequences and the reference video sequence, respectively. It can be seen that in Eq. \ref{eq:DegradationPenalty}, $PD_i$ is positive correlated to the difference between $A_i$ and $A_\mathrm{ref}$. In other words, less penalty will be assigned if $A_i$ is more closer to $A_\mathrm{ref}$. $PD_i$ ranges from $0$ to $1$. If $A_i$ is much greater than $A_\mathrm{ref}$ (e.g., $A_i>2A_\mathrm{ref}$) which rarely happens in the robustness test, we limit the penalty to be $1$. Actually, the penalty of accuracy degradation $PD_i$ describe the invariance property of the detection accuracy.

Furthermore, the non-monotonicity penalty of the $i^{th}$ distorted sequence $PM_i$ is formulated as:
\begin{equation}\label{eq:MonotonicityPenalty}
  PM_i =
   \begin{cases}
   0 ~&A_{i} \leq A_{i-}\\
   \min\{1,(\frac{A_i-A_{i-}}{A_{i-}})^2\} ~&A_{i}>A_{i-},
   \end{cases}
\end{equation}
where $A_{i}$, $A_{i-}$, and $A_\mathrm{ref}$ are the detection accuracy on the $i^{th}$, $i-^{th}$ distorted video sequences and the reference sequence, respectively.

Here, we give the definition of the $i-{th}$ distorted video sequence. With distortion type of $\mathrm{qp}$, $\mathrm{res}$ or $\mathrm{wn}$, we simply rank the distorted sequences with quality descending. The $(i-)^{th}$ distorted sequence in Eq. \ref{eq:MonotonicityPenalty} is just next to the $i^{th}$ one and with better quality\footnote{If $i=1$, $(i-)^{th}$ is the reference video sequence.}. With distortion type of $\mathrm{bv}$, the distorted sequences are divided into high brightness sequences and low brightness sequences comparing with the reference video. By ranking these two groups separately with quality descending, the $(i-)^{th}$ distorted sequence is next to the $i^{th}$ one in the same group with better quality.

To meet the monotonicity criterion, $A_{i}$ is supposed to be not larger than $A_{i-}$ since the quality of the $i^{th}$ distorted video sequence is worse than that of the $(i-)^{th}$ distorted video sequence. Thus, if $A_{i}>A_{i-}$, penalty will be assigned as shown in Eq. \ref{eq:MonotonicityPenalty}. With the same concern in Eq. \ref{eq:DegradationPenalty}, if $A_i$ is much greater than $A_\mathrm{i-}$ (e.g., $A_i>2A_\mathrm{i-}$), we limit the penalty to be $1$.

\begin{figure*}[!htbp]
\setlength{\abovecaptionskip}{2pt}
  \centering
   \includegraphics[width=0.96\textwidth]{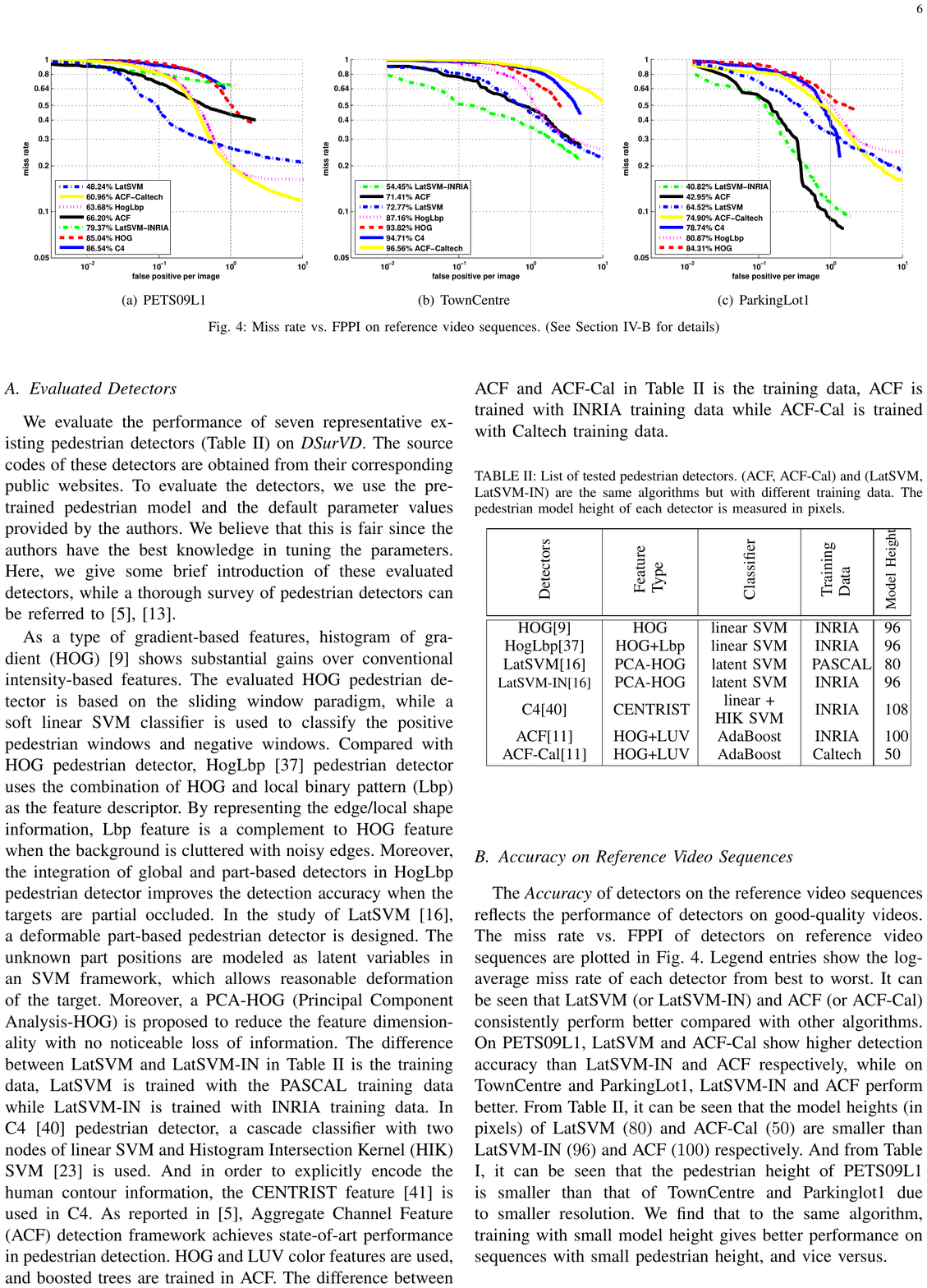}%
  \caption{Miss rate vs. FPPI on reference video sequences. (See Section \ref{subsec:RefAccuracy} for details)}\label{fig:AccuracyPlot}
\end{figure*}

Based on these two penalty functions, we compute the \emph{Stability} with given distortion type $\mathrm{x}$ (e.g., $\mathrm{x}$ can be $\mathrm{qp}$, $\mathrm{res}$, $\mathrm{wn}$ and $\mathrm{bv}$) as:
\begin{equation}\label{eq:A2S}
    \mathbf{S}.\mathrm{x} = 1-\sqrt{\frac{1}{N_\mathrm{x}}\sum\limits_{i=1}^{N_\mathrm{x}}\omega PD_i+(1-\omega)PM_i},
\end{equation}
where $\omega$ is the weighting parameter between $PD_i$ and $PM_i$, which is used to adjust the importance of these two factors, and it ranges from $0$ to $1$.

Fig. \ref{fig:Accuracy2Stability} shows the flowchart of computing \emph{Stability}. The quantified \emph{Stability} ranges from $0$ to $1$, and a higher value shows more stable performance of a pedestrian detector. The stability value of an ideal stable detector should be $1$ with $A_i \triangleq A_\mathrm{ref}$.

\section{Evaluation Result}
\label{sec:EvalResults}
In this section, we show the evaluation result of several existing pedestrian detectors on \emph{DSurVD}\footnote{Only results on sequences PETS09L1, TownCentre and ParkingLot1 are shown in this paper due to the page limitation. The full results can be achieved on: \url{https://sites.google.com/site/sorsyuanyuan/home/rdetection}}. The comparison and robustness between the tested detectors are given, along with the discussion about the evaluation result.

\subsection{Evaluated Detectors}
We evaluate the performance of seven representative existing pedestrian detectors (Table \ref{tab:Algorithms}) on \emph{DSurVD}. The source codes of these detectors are obtained from their corresponding public websites. To evaluate the detectors, we use the pre-trained pedestrian model and the default parameter values provided by the authors. We believe that this is fair since the authors have the best knowledge in tuning the parameters. Here, we give some brief introduction of these evaluated detectors, while a thorough survey of pedestrian detectors can be referred to \cite{Benenson2014Eccvw,enzweiler2009monocular}.

As a type of gradient-based features, histogram of gradient (HOG) \cite{dalal2005histograms} shows substantial gains over conventional intensity-based features. The evaluated HOG pedestrian detector is based on the sliding window paradigm, while a soft linear SVM classifier is used to classify the positive pedestrian windows and negative windows. Compared with HOG pedestrian detector, HogLbp \cite{wang2009hoglbp} pedestrian detector uses the combination of HOG and local binary pattern (Lbp) as the feature descriptor. By representing the edge/local shape information, Lbp feature is a complement to HOG feature when the background is cluttered with noisy edges. Moreover, the integration of global and part-based detectors in HogLbp pedestrian detector improves the detection accuracy when the targets are partial occluded. In the study of LatSVM \cite{felzenszwalb2010object}, a deformable part-based pedestrian detector is designed. The unknown part positions are modeled as latent variables in an SVM framework, which allows reasonable deformation of the target. Moreover, a PCA-HOG (Principal Component Analysis-HOG) is proposed to reduce the feature dimensionality with no noticeable loss of information. The difference between LatSVM and LatSVM-IN in Table \ref{tab:Algorithms} is the training data, LatSVM is trained with the PASCAL training data while LatSVM-IN is trained with INRIA training data. In C4 \cite{wu2011real} pedestrian detector, a cascade classifier with two nodes of linear SVM and Histogram Intersection Kernel (HIK) SVM \cite{maji2008classification} is used. And in order to explicitly encode the human contour information, the CENTRIST feature \cite{wu2011centrist} is used in C4. As reported in \cite{Benenson2014Eccvw}, Aggregate Channel Feature (ACF) detection framework achieves state-of-art performance in pedestrian detection. HOG and LUV color features are used, and boosted trees are trained in ACF. The difference between ACF and ACF-Cal in Table \ref{tab:Algorithms} is the training data, ACF is trained with INRIA training data while ACF-Cal is trained with Caltech training data.

\begin{table}[!htbp] \centering \small
\caption{List of tested pedestrian detectors. (ACF, ACF-Cal) and (LatSVM, LatSVM-IN) are the same algorithms but with different training data. The pedestrian model height of each detector is measured in pixels.}
  \label{tab:Algorithms} 
\begin{tabular}{|m{1.85cm}<{\centering}|m{1.45cm}<{\centering}|m{1.6cm}<{\centering}|m{1.0cm}<{\centering}|m{0.3cm}<{\centering}|}
  \hline
  \rotatebox{90}{Detectors} & \rotatebox{90}{\minitab[l]{Feature\\ Type}} & \rotatebox{90}{Classifier} & \rotatebox{90}{\minitab[l]{Training\\ Data}} & \rotatebox{90}{\footnotesize Model Height} \\
  \hline
  \hline
  HOG\cite{dalal2005histograms} & HOG & linear SVM & INRIA & 96 \\
  HogLbp\cite{wang2009hoglbp} & HOG+Lbp & linear SVM & INRIA & 96 \\
  LatSVM\cite{felzenszwalb2010object} & PCA-HOG & latent SVM & PASCAL & 80 \\
  \footnotesize LatSVM-IN\cite{felzenszwalb2010object} & PCA-HOG & latent SVM & INRIA & 96 \\
  C4\cite{wu2011real} & CENTRIST & linear + HIK SVM & INRIA & 108 \\
  ACF\cite{dollar2014ACF} & HOG+LUV & AdaBoost & INRIA & 100 \\
  ACF-Cal\cite{dollar2014ACF} & HOG+LUV & AdaBoost & Caltech & 50 \\
  \hline
\end{tabular}
\end{table}

\subsection{Accuracy on Reference Video Sequences}
\label{subsec:RefAccuracy}

The \emph{Accuracy} of detectors on the reference video sequences reflects the performance of detectors on good-quality videos. The miss rate vs. FPPI of detectors on reference video sequences are plotted in Fig. \ref{fig:AccuracyPlot}. Legend entries show the log-average miss rate of each detector from best to worst. It can be seen that LatSVM (or LatSVM-IN) and ACF (or ACF-Cal) consistently perform better compared with other algorithms. On PETS09L1, LatSVM and ACF-Cal show higher detection accuracy than LatSVM-IN and ACF respectively, while on TownCentre and ParkingLot1, LatSVM-IN and ACF perform better. From Table \ref{tab:Algorithms}, it can be seen that the model heights (in pixels) of LatSVM ($80$) and ACF-Cal ($50$) are smaller than LatSVM-IN ($96$) and ACF ($100$) respectively. And from Table \ref{tab:SeqStat}, it can be seen that the pedestrian height of PETS09L1 is smaller than that of TownCentre and Parkinglot1 due to smaller resolution. We find that to the same algorithm, training with small model height gives better performance on sequences with small pedestrian height, and vice versus.
\\

\subsection{Quadrangle: Robustness Representation}
\label{subsec:RobustnessQuad}
\begin{figure*}[!htbp]
\setlength{\abovecaptionskip}{2pt}
  \centering

   \includegraphics[width=0.96\textwidth]{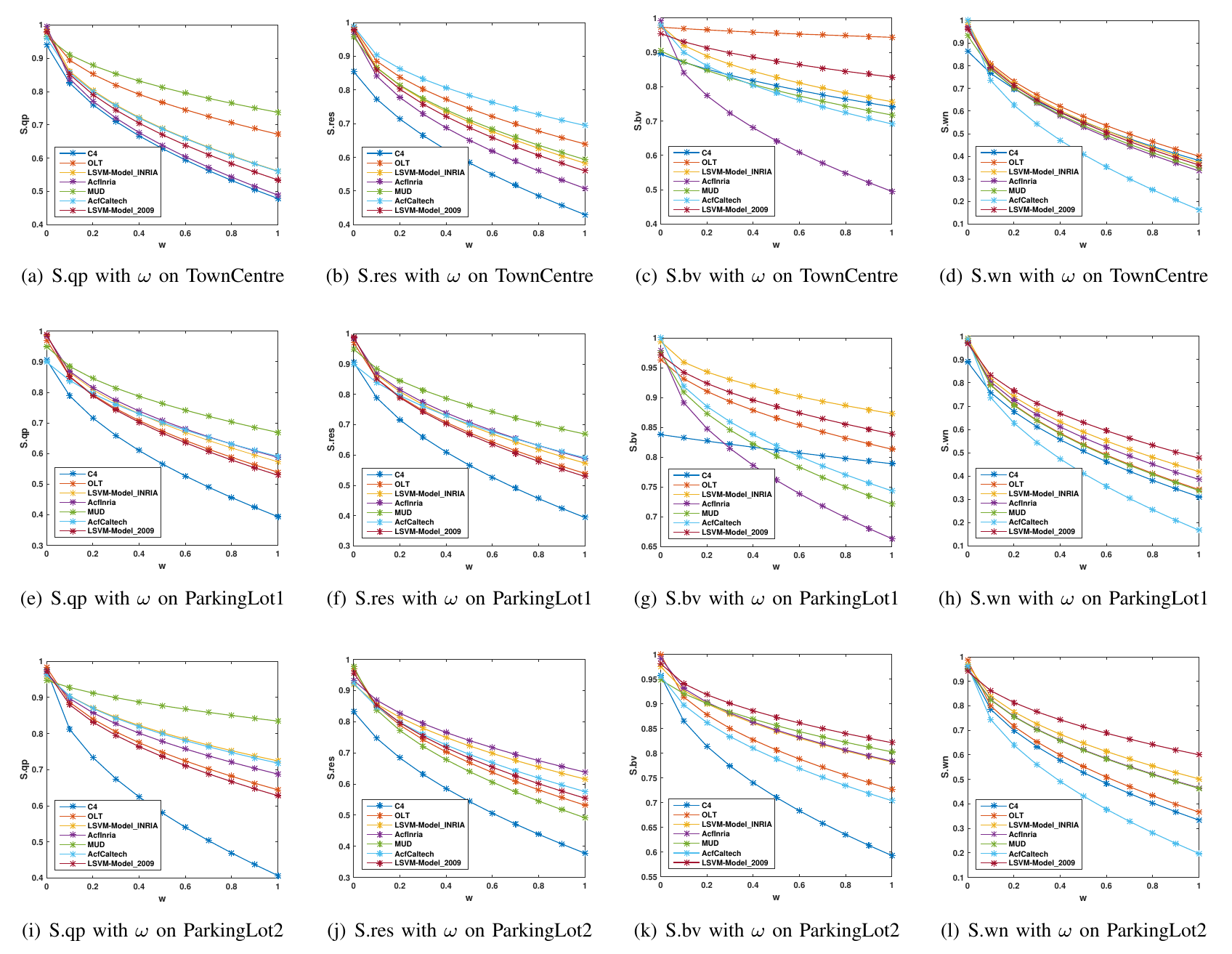}%

  \caption{The performance of \emph{Stability} $S$ with different weighting parameters $\omega$ on different datasets. X-axis refers to the different weighting parameters $\omega$ in the range from 0 to 1, and y-axis represents the performance of \emph{Stability} $S$ with four attributions (S.qp: Quantization Parameters, S.res: Resolution, S.bv: Brightness Variation, S.wn: White Noise).}\label{fig:StabilityPlot1}
\end{figure*}

With the definition in Sec. \ref{sec:AccySta}, the robustness of a detector can be described by a combination of $Accuracy$ ($A_\mathrm{ref}$) on good-quality video and $Stability$ ($\mathbf{S}$) with four types of distortions. The detection accuracy of detectors on distorted video sequences are computed and plotted in Fig. \ref{fig:StabilityPlot}. Besides, the performance of $Stability$ with different weighting parameters $\omega$ are plotted in Fig. \ref{fig:StabilityPlot1}. From Fig. \ref{fig:StabilityPlot1} and Fig. \ref{fig:StabilityPlot}, it can be seen that most detectors follow the monotonicity criterion when the video quality drops. Additionally, with different weighting parameters $\omega$, the ranking order of $Stability$ $S$ from different pedestrian detectors are almost kept stable, which demonstrates that the adjustment of parameter $\omega$ has little effect on the ranking order of $Stability$ $S$ by different pedestrian detection algorithms. In other words, with larger parameter $\omega$, the performance of the pedestrian detection methods decreases when the video quality drops. However, there might be some pedestrian detector with bad monotonicity in the literature. Thus, we hold the second factor with small value to provide good extensibility for the proposed metric. This is the reason why we assign more weighting to the penalty of accuracy degradation $PD$ by setting w=0.8 in Eq. (\ref{eq:A2S}) when computing the detection performance \emph{Stability}. In the experiment, we found that the detection accuracy would not decrease greatly with video quality dropping on a robust pedestrian detector, which demonstrates that the initial hypothesis of monotonicity is convincing. In the future, we will further investigate into this weighting parameter to design better metric.

When comparing the robustness of two detectors, we first consider the value of $A_\mathrm{ref}$. If there exists a large $A_\mathrm{ref}$ difference between the compared detectors, the detector with the larger $A_\mathrm{ref}$ is more preferable and we should take less consideration on the $Stability$. On the other hand, if the compared detectors are with a similar value of $A_\mathrm{ref}$ (this is likely the case for the relevant  state-of-the-art detectors), then $\mathbf{S}$ becomes an important criterion to robustness.

Based on this analysis, we propose the robustness quadrangle (as shown in Fig. \ref{fig:RobustnessQuad}) to visualize $A_\mathrm{ref}$ and $\mathbf{S}$. For each quadrangle, the heights of four angles represent the $Stability$ to the four types of distortion respectively ($\mathbf{S}.\mathrm{qp}, \mathbf{S}.\mathrm{res}, \mathbf{S}.\mathrm{wn}, \mathbf{S}.\mathrm{bv}$). The center point of each quadrangle indicates $A_\mathrm{ref}*\lambda$ where $\lambda$ is a scaling factor of $A_\mathrm{ref}$ in robustness and decides the range of x-axis of the robustness quadrangles figure.

By given a non-zero value to $\lambda$ (e.g. $\lambda=5$, $\lambda=2$), the robustness quadrangles of evaluated detectors are shown in the left and middle column of Fig. \ref{fig:RobustnessQuad}. If two detectors are with large $A_\mathrm{ref}$ difference, we can intuitively read the $A_\mathrm{ref}$ difference based on the distance between two robustness quadrangles. If the $A_\mathrm{ref}$ values of two detectors are similar, the center points of two quadrangles are close and we can straightforward compare their $\mathbf{S}$ values by the corners of two overlapped quadrangles. The red square on the most right hand side with dashed boundary represents the ideal detector whose $A_\mathrm{ref}=1$ and $\mathbf{S} = [1,1,1,1]$.
%
%

If we want to emphasis $A_\mathrm{ref}$ more in the robustness quadrangle figure, we can set larger $\lambda$ (e.g. $\lambda=5$, the left column of Fig. \ref{fig:RobustnessQuad}). Thus the differences of $A_\mathrm{ref}$ between detectors will be amplified on the x-axis. If we want to emphasis $\mathbf{S}$ more in the robustness quadrangle figure, we can set smaller $\lambda$ with the same reason (e.g. $\lambda=2$, the middle column of Fig. \ref{fig:RobustnessQuad}). $\lambda=0$ is an extreme case that we only compare $\mathbf{S}$ of detectors while all the center points of quadrangles converge to $[0,0]$ and we cannot see any difference between $A_\mathrm{ref}$. The right column of Fig \ref{fig:RobustnessQuad} shows the the robustness quadrangles with $\lambda=0$.

\subsection{Stability with QP Variation}
\label{subsec:StabilityQP}
The first row of Fig. \ref{fig:StabilityPlot} shows the $Accuracy$ vs. PSNR curves of evaluated detectors with three different scenes (Campus, Town Centre and Car Parking). It can be seen that, in general, the $Accuracy$-PSNR curves of most detectors are monotonic. And the degradation of $Accuracy$ of detectors is not obvious before some critical PSNR points (e.g., before $35$dB). One noticeable fact is that, to each detector, the $Accuracy$ fluctuates around $A_\mathrm{ref}$ before it dramatically decreases. This is because when the pixel values slightly changes due to compression, it would affect the decision of the algorithms on detections near to the threshold.

From the $Stability$ values of left-upper corners in Fig. \ref{fig:RobustnessQuad} and the ranking in Fig. 6(a), Fig. 6(b), Fig. 6(c), HogLbp is most stable with QP variation among the evaluated detectors by ranking 1st on both TownCentre and ParkingLot1 and 2nd on PETS09L1. Comparing to HOG, the main modification of HogLbp is the feature type. Hence, Lbp feature \cite{ojala1996comparative} can be an important complement to HOG feature in pedestrian detection on heavily compressed surveillance videos with large QP. Moreover, with QP variation, ACF-Cal shows higher stability than ACF, and LatSVM-IN shows higher stability than LatSVM. This indicates that the training data is another important factor to detection stability. With the same algorithm of ACF, the detector trained with Caltech \cite{dollar2012pedestrian} data set performs more stable than the detector trained with INRIA \cite{dalal2005histograms} data set; with the same algorithm of LatSVM, the detector trained with INRIA data set performs more stable than the detector trained with PASCAL \cite{everingham2010pascal}. More exploration is needed toward generalization of learning-based approaches and tackle overfitting.

\subsection{Stability with Resolution Variation}
\label{subsec:StabilityRes}
To resolution degradation of video sequences, pedestrian height change has the most direct impact which affects the detection accuracy. We plot the $Accuracy$ vs. mean Pedestrian Height (in pixels) curves of evaluated detectors in the second row of Fig \ref{fig:StabilityPlot}. It can be seen in most of the cases, videos with larger pedestrian height are with higher $Accuracy$ values. Also, before the pedestrian height decreases to some critical points (e.g., $40$ to PETS09L1, and $60$ to TownCentre and ParkingLot1), the $Accuracy$ values of detectors are relatively stable with pedestrian height change.

From the $Stability$ values of right-upper corners in Fig. \ref{fig:RobustnessQuad} and the ranking in Fig. 6(d), Fig. 6(e), Fig. 6(f), we can see that ACF-Cal performs more constantly even on videos with low pedestrian height. Note that, the ACF-Cal is with the lowest model height (50) for training among the evaluated detectors. It reveals that the lower model height for training can result in stable performance to videos with low pedestrian height (or low resolution). However, a detector with too low model height may cause low $Accuracy$ in detection as well (see the performance of ACF-Cal on TownCentre). We should be careful in defining the model height in the later detector designing to achieve both high $Accuracy$ and $Stability$.

\subsection{Stability with Additive White Noise}
\label{subsec:StabilityWN}
The third row of Fig. \ref{fig:StabilityPlot} shows the $Accuracy$ vs. PSNR curves of evaluated detectors with different levels of additive white noise. It can be seen that, for most detectors, there is no obvious degradation of the $Accuracy$ before some critical PSNR (e.g., before $40$dB for the cases being studied) points. For noisy sequences of PETS09L1 and ParkingLot1, LatSVM and LatSVM-IN shows preferable stability comparing with other detectors based on the $Stability$ values of right-bottom corners in Fig. \ref{fig:RobustnessQuad} and ranking in Fig. 6(g), Fig. 6(h), Fig. 6(i). For noisy sequences of TownCentre, LatSVM and LatSVM-IN rank the third and fourth just behind HOG and C4. However, the high $Stability$ values of HOG and C4 on TownCentre are meaningless, due to their low detection $Accuracy$ on the reference video of TownCentre (less than 0.1).

It has been demonstrated that, the tail of singular values are dominated by the white Gaussian noise while the first few singular values are dominated by the image structure when performing SVD to noisy images \cite{liu2013additive,narwaria2012svd}. Likewise, the PCA-HOG used in LatSVM and LatSVM-IN can be a good way for increasing the detection accuracy on noisy video sequences, by only using the principle components which are more robust to white noise.

\subsection{Stability with Brightness Variation}
\label{subsec:StabilityBV}
\begin{figure*}[!htbp]
\setlength{\abovecaptionskip}{2pt}
  \centering

   \includegraphics[width=0.96\textwidth]{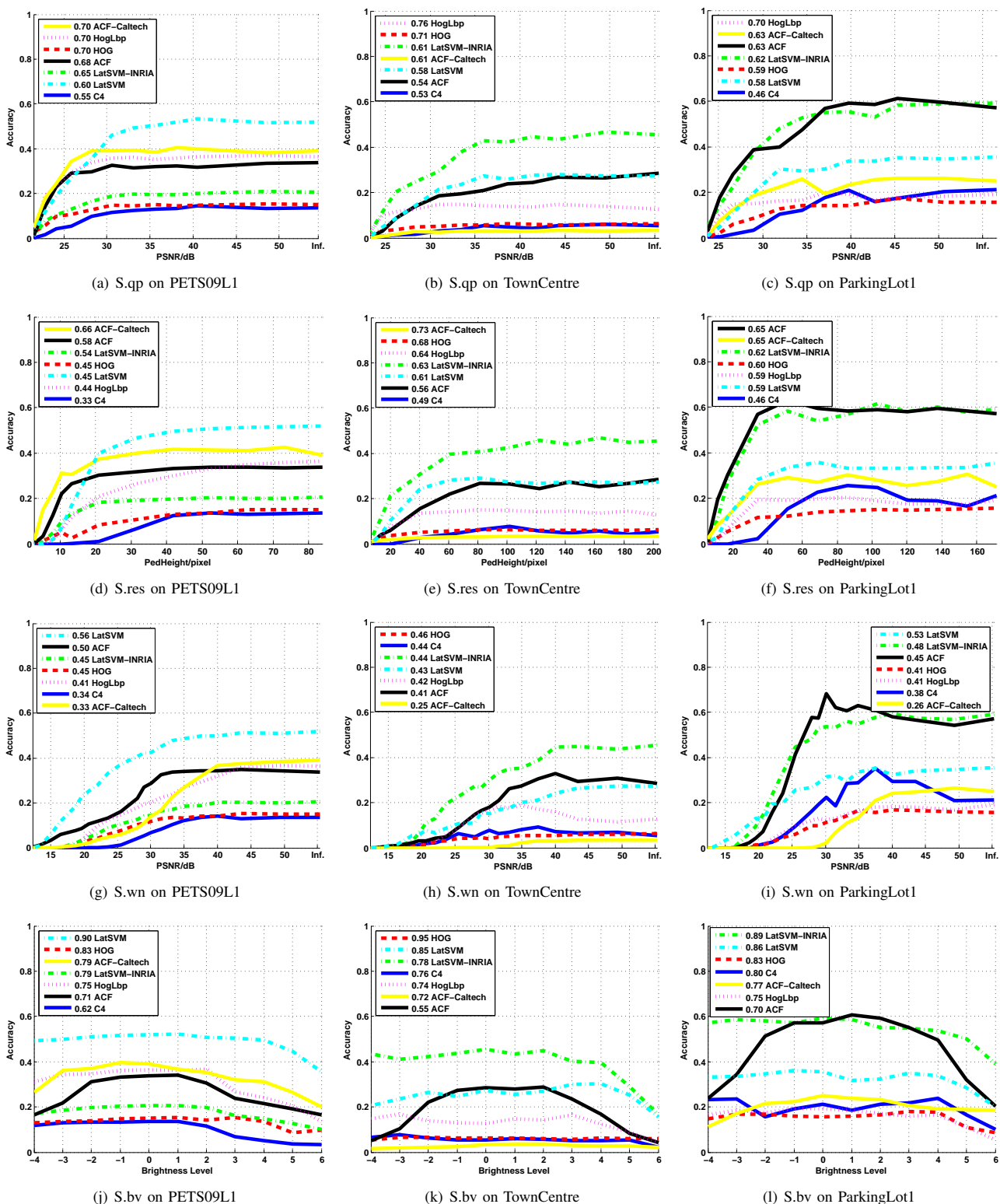}%

  \caption{Detection accuracy variation of evaluated detectors with different types of distortion. In the first row and third row, PSNR=Inf. refers to the reference video sequences with the best quality. In the second row, the x-axis refers to the average pedestrian heights of the tested sequences. In the fourth row, the x-axis indicates the brightness level. ``$0$'' denotes the reference video sequences while negative values denote low brightness sequences and positive values denote high brightness sequences.}\label{fig:StabilityPlot}
\end{figure*}

The last row of Fig. \ref{fig:StabilityPlot} shows the $Accuracy$ changing to brightness variation of the evaluated detectors. ``$0$'' on brightness level axis denotes the reference video sequences, while negative values indicate low-brightness sequences and positive values indicate high-brightness sequences.

\begin{landscape}

\begin{figure}[!htbp]
\setlength{\abovecaptionskip}{2pt}
\centering

  \subfigure[PETS09L1,$\lambda = 5$]{\label{fig:subfig:QuadPETS09L1}
   \includegraphics[width=0.65\textwidth]{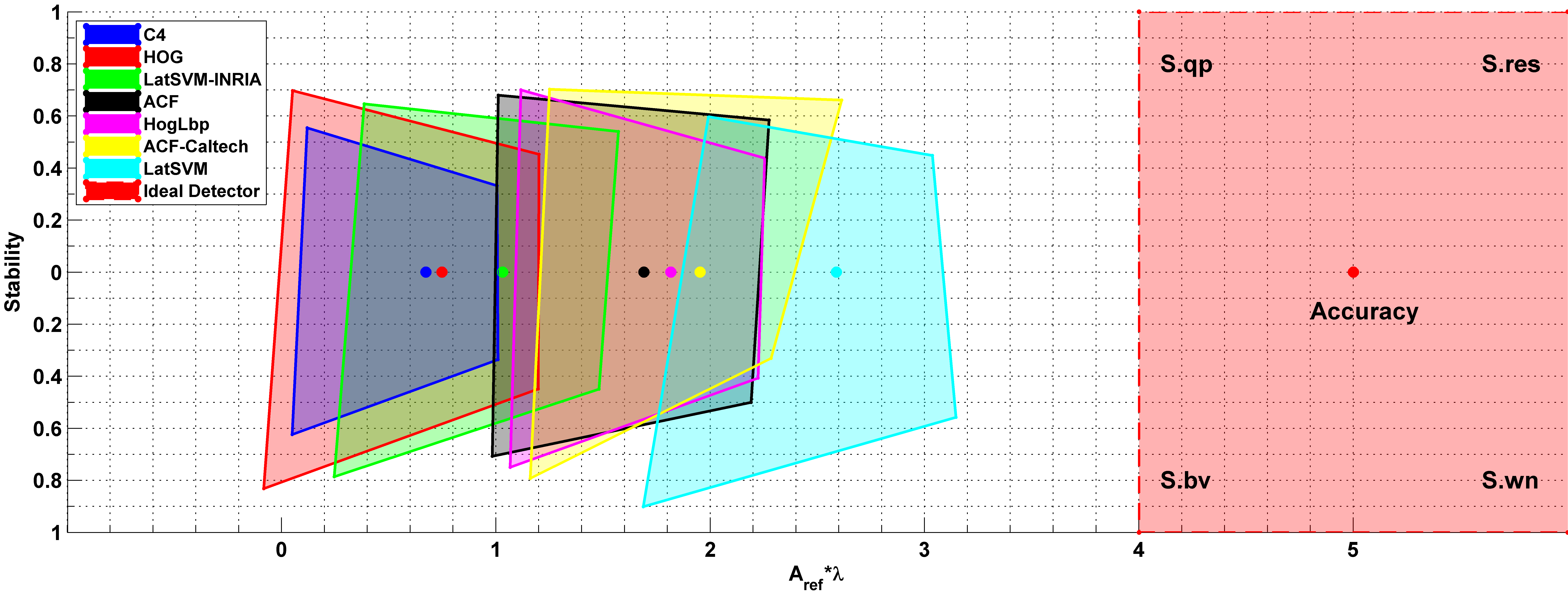}}%
  \subfigure[PETS09L1,$\lambda = 2$]{\label{fig:subfig:QuadPETS09L1Lambda2}
   \includegraphics[width=0.43\textwidth]{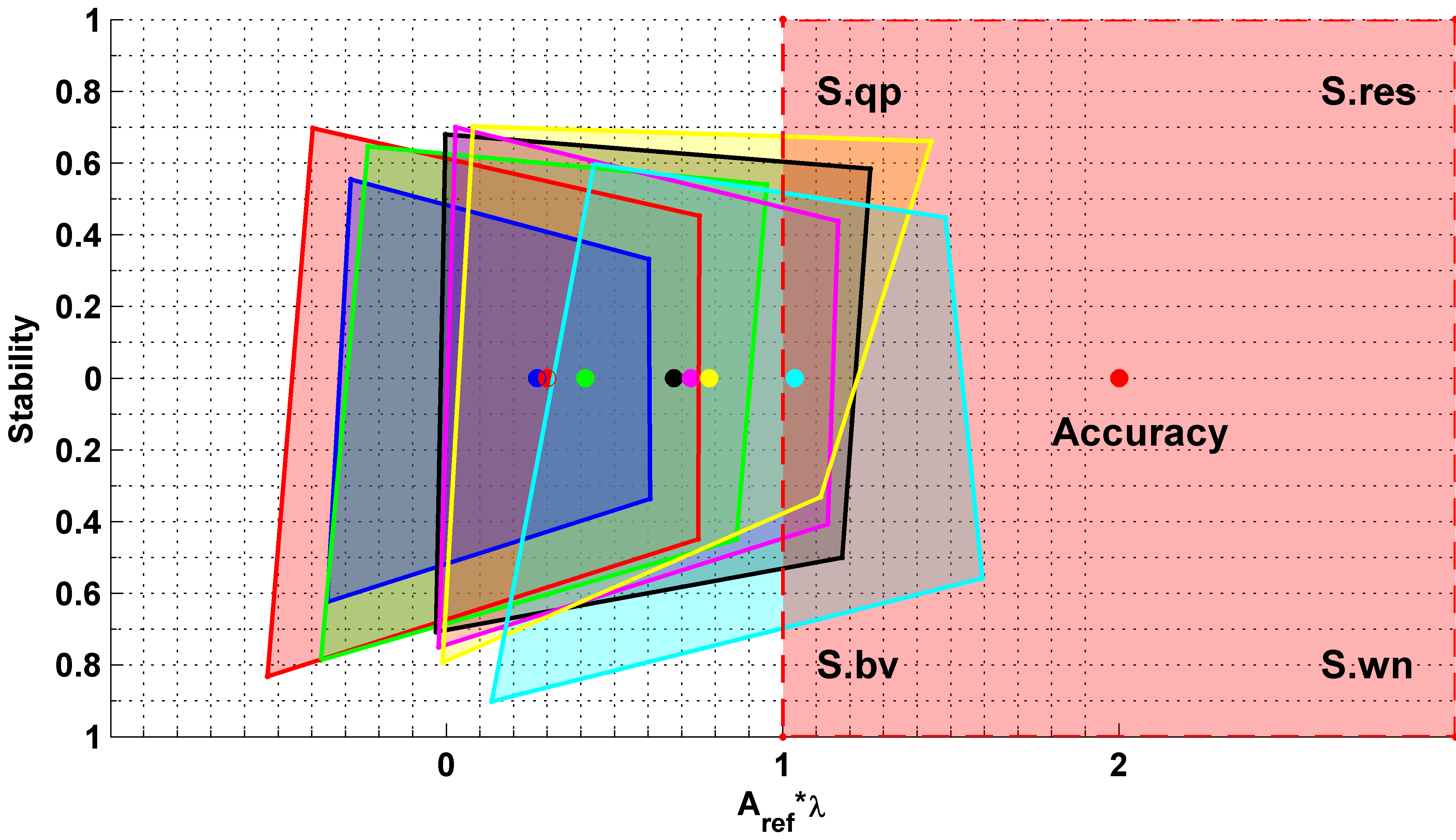}}%
  \subfigure[PETS09L1,$\lambda = 0$]{\label{fig:subfig:QuadPETS09L1Lambda0}
   \includegraphics[width=0.245\textwidth]{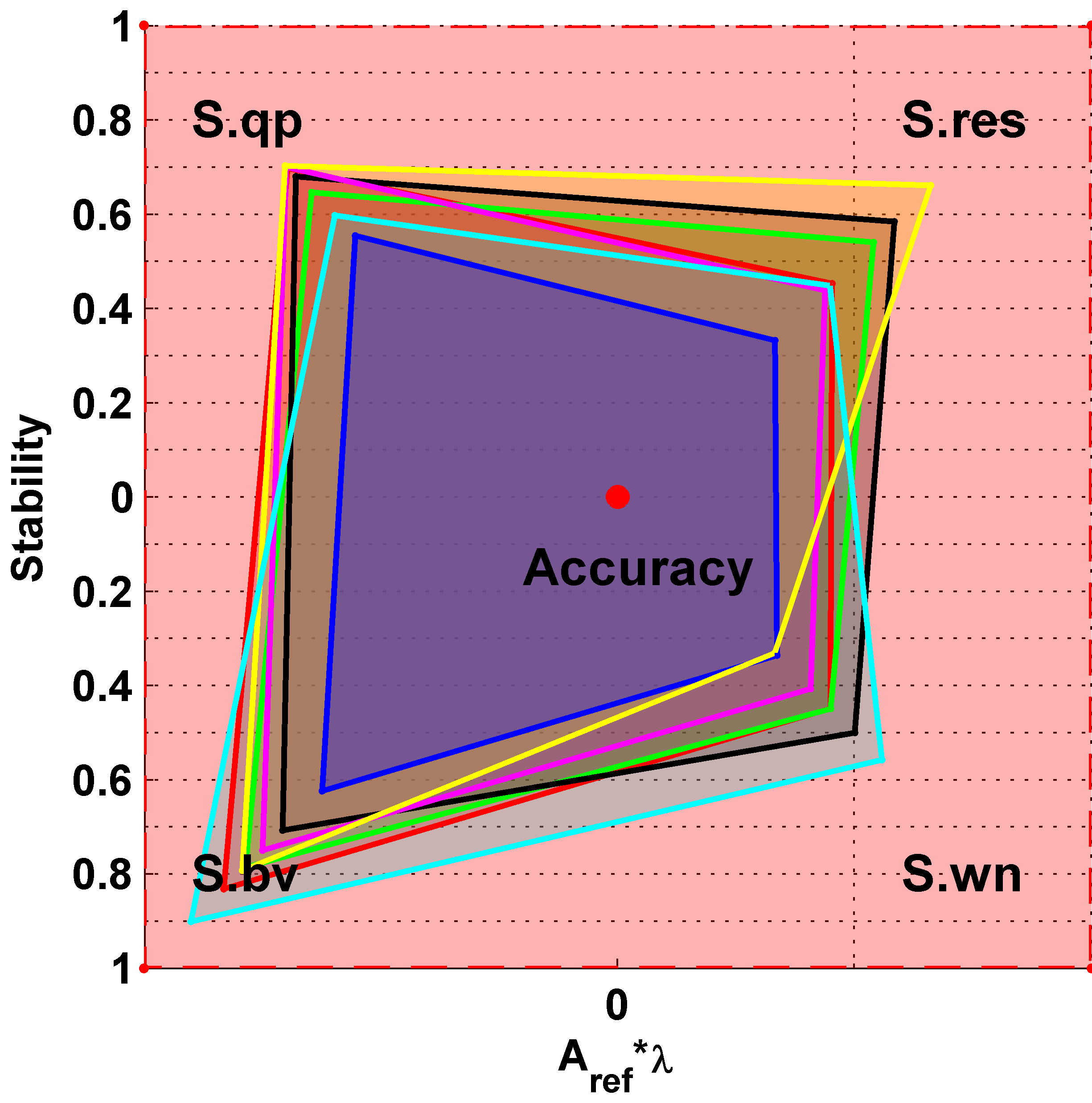}}%

  \subfigure[TownCentre,$\lambda = 5$]{\label{fig:subfig:QuadTownCentre}
   \includegraphics[width=0.65\textwidth]{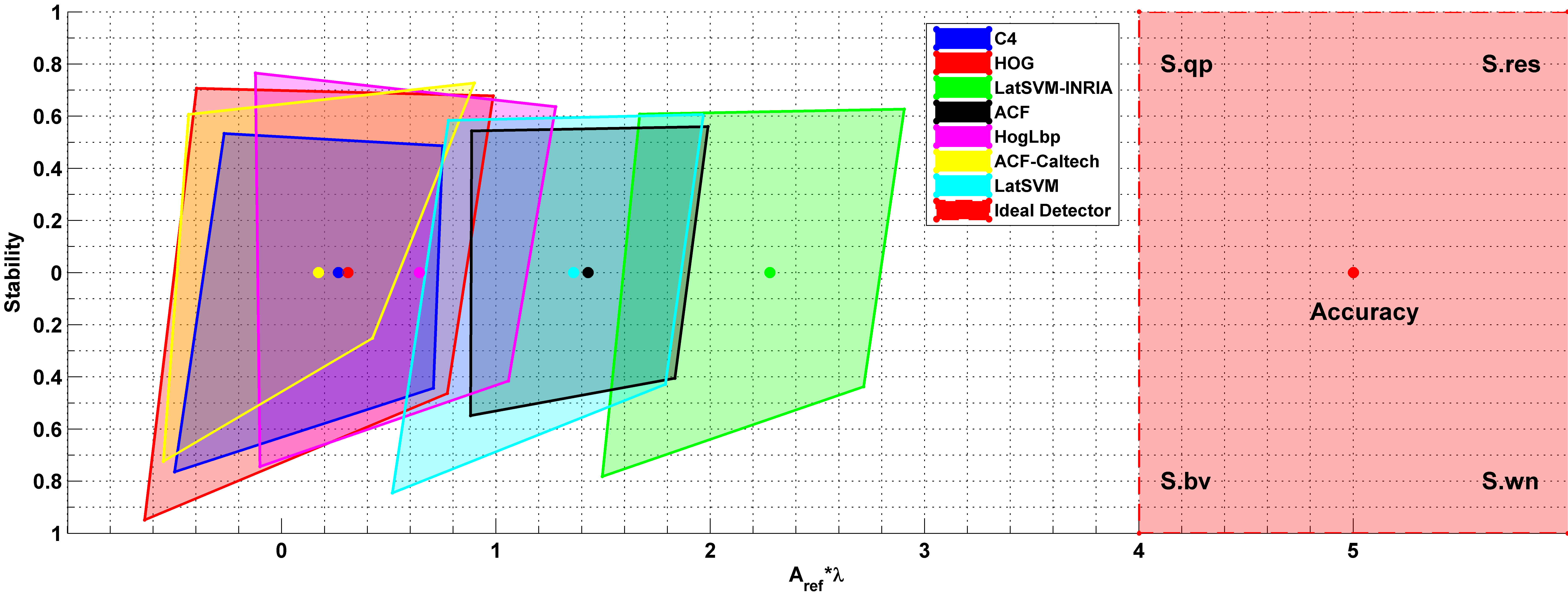}}%
  \subfigure[TownCentre,$\lambda = 2$]{\label{fig:subfig:QuadTownCentreLambda2}
   \includegraphics[width=0.43\textwidth]{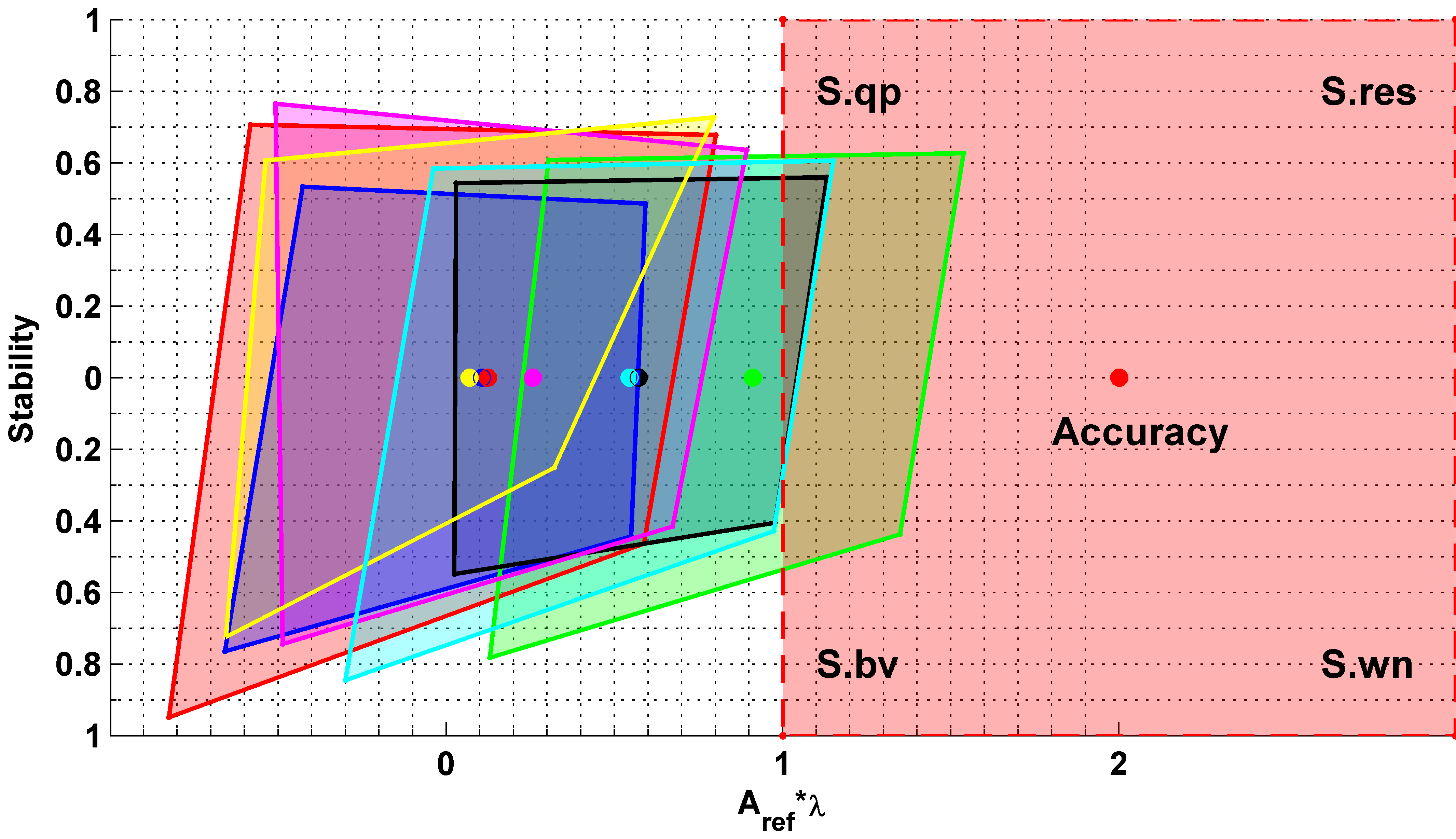}}%
  \subfigure[TownCentre,$\lambda = 0$]{\label{fig:subfig:QuadTownCentreLambda0}
   \includegraphics[width=0.245\textwidth]{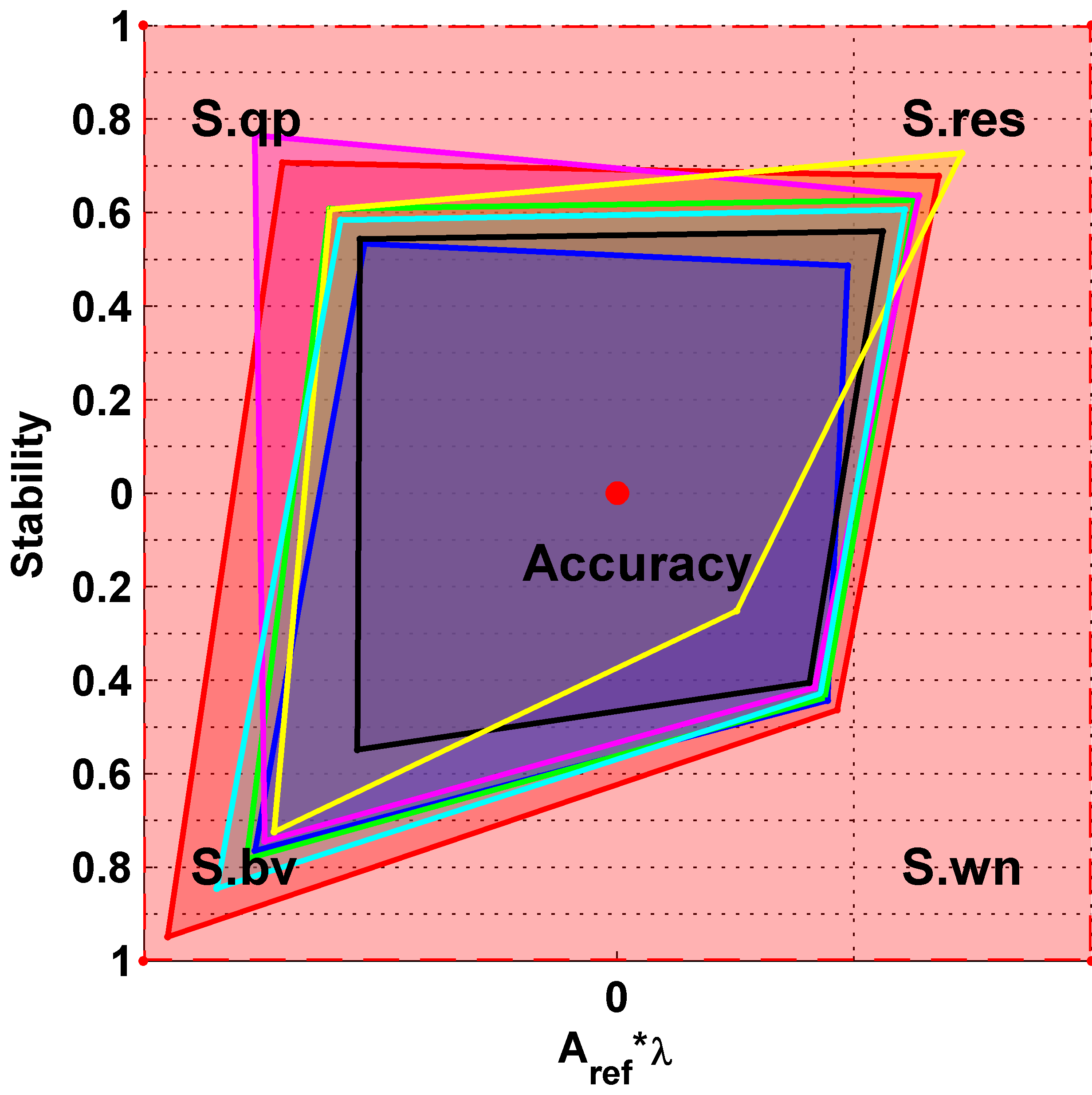}}%

  \subfigure[ParkingLot1,$\lambda = 5$]{\label{fig:subfig:QuadParkingLot1}
   \includegraphics[width=0.65\textwidth]{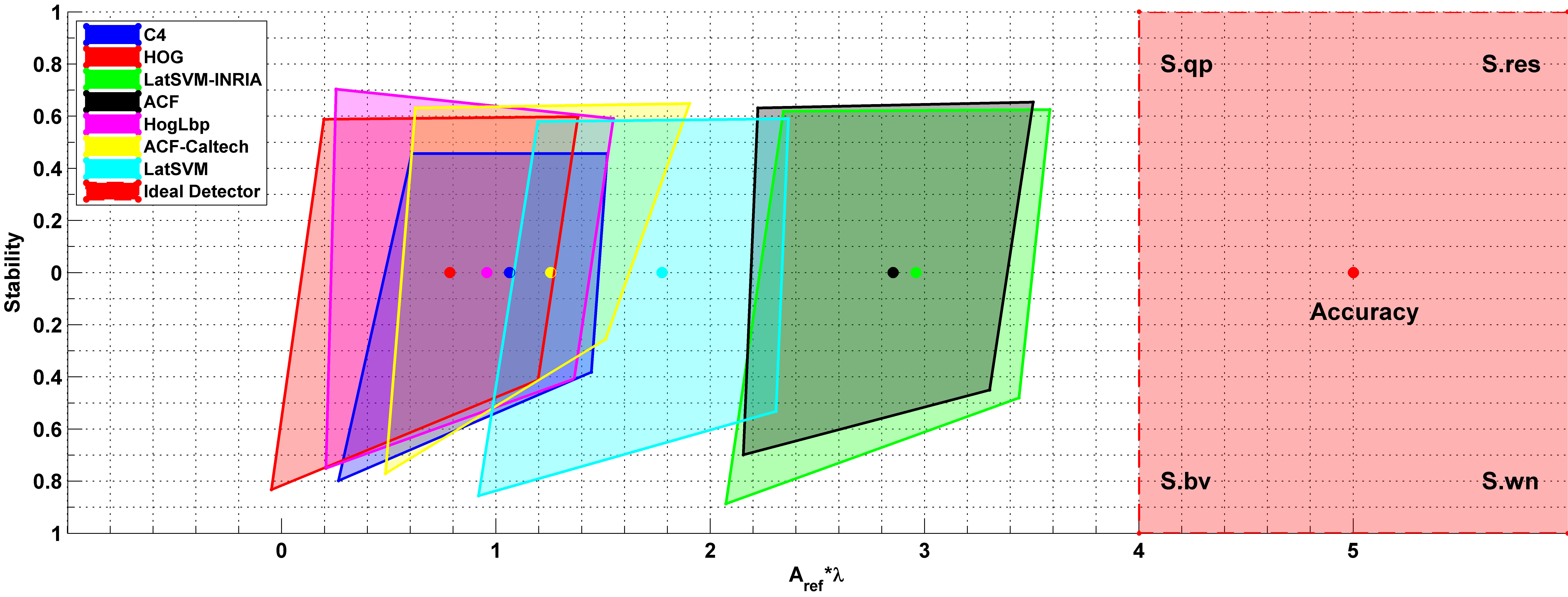}}%
  \subfigure[ParkingLot1,$\lambda = 2$]{\label{fig:subfig:QuadParkingLot1Lambda2}
   \includegraphics[width=0.43\textwidth]{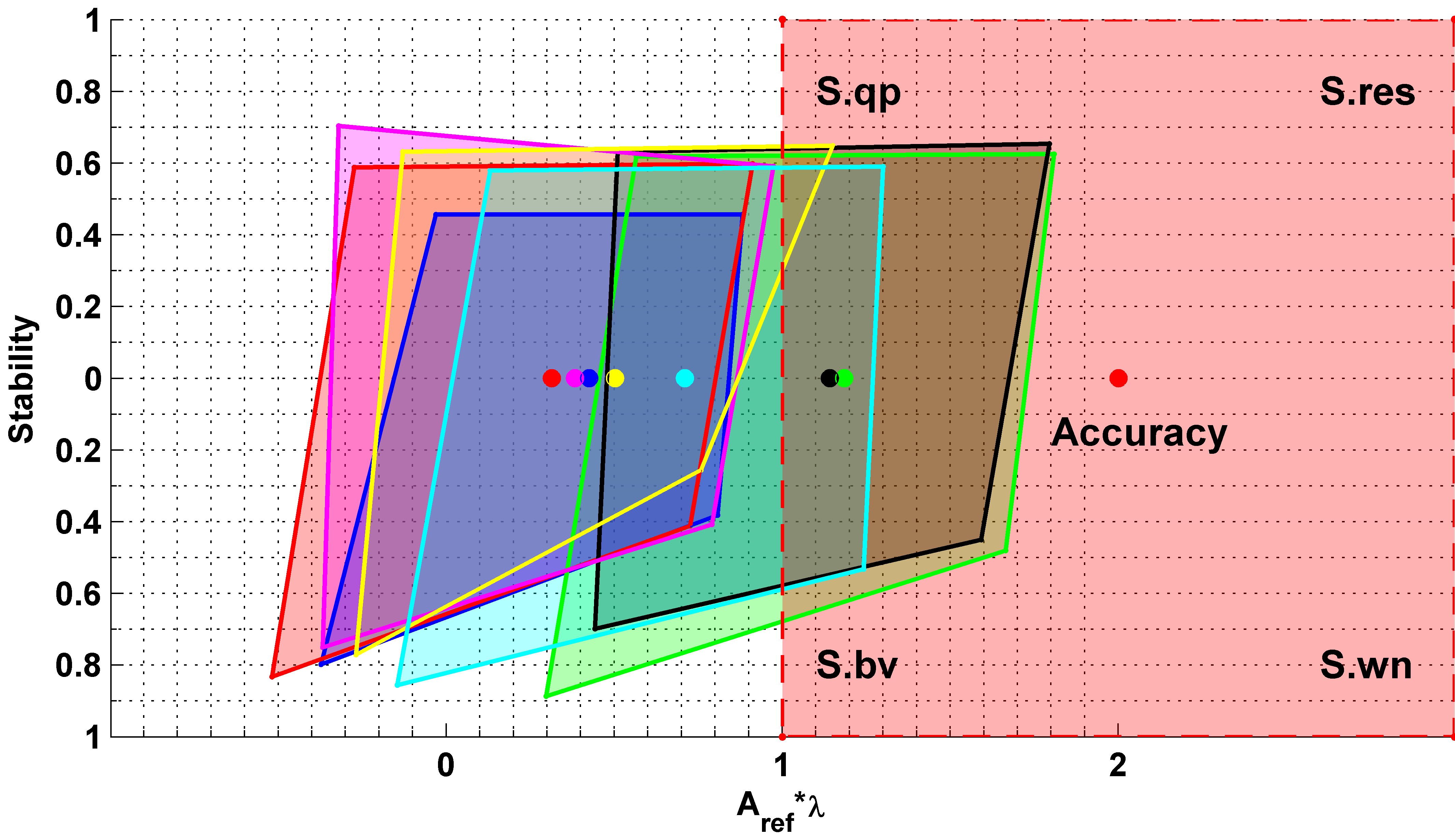}}%
  \subfigure[ParkingLot1,$\lambda = 0$]{\label{fig:subfig:QuadParkingLot1Lambda0}
   \includegraphics[width=0.245\textwidth]{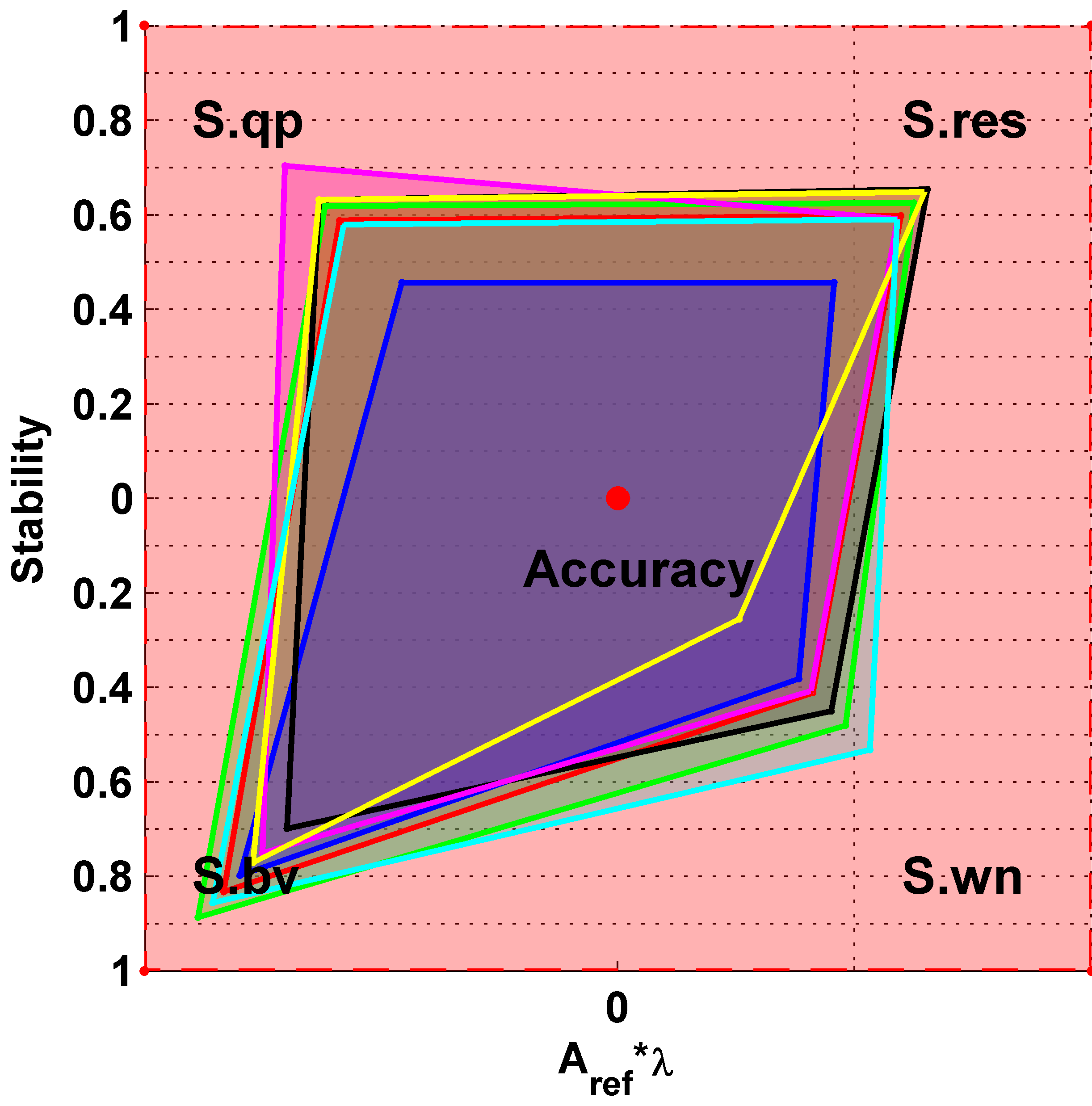}}%
  \caption{Robustness Quadrangles on tested sequences. The center location of each quadrangle read from x-axis denotes $A_\mathrm{ref}*\lambda$ of each detector. The heights of four corners read from y-axis denote stability values with four types of distortions, respectively (QP variation, Resolution variation, White noise and Brightness variation from left-upper corner to left-bottom corner clockwise). For the ideal detector which plotted in red color with dashed line, the center of corresponded quadrangle is $1*\lambda$ and the heights of all the four corners are $1$. When comparing the robustness of two detectors, the center locations of quadrangles are first compared and quadrangle on the right side is preferred. When the center locations of two quadrangles are close, the differences of stability values with different distortion types can be easily read from the corner heights. $\lambda$ is a scaling factor of $A_\mathrm{ref}$ in robustness measurement, by setting $\lambda=0$ (shown in the right column), only the differences of stability values can be read and the differences between $A_\mathrm{ref}$ are obscured.(See Section \ref{subsec:RobustnessQuad} for details)}\label{fig:RobustnessQuad}
\end{figure}

\end{landscape}

From the $Stability$ values of left-bottom corners in Fig. \ref{fig:RobustnessQuad} and ranking in Fig. 6(j), Fig. 6(k), Fig. 6(l), it can be seen that detectors with HOG based features alone (HOG, LatSVM and LatSVM-IN) are more stable with brightness variation comparing with other evaluated detectors, in line with the claim in \cite{dalal2005histograms} that HOG feature is invariant to illumination variation. Moreover, it can be seen that ACF and ACF-Cal which use LUV feature channel are more sensitive to brightness variation. Another interest finding is that, evaluated detectors shows higher stability with brightness variation than QP, resolution variation and additive white noise. One possibility can be that, brightness variation caused by environment illumination change has been widely studied by the community of pedestrian detection, and thus the brightness variation problem is better addressed in recent pedestrian detection studies.

\section{Summary and Discussion}
\label{sec:Discussion}
In this paper, we have introduced the \emph{DSurVD} for evaluating the robustness of pedestrian detectors to video distortions including H.264 compression distortion, resolution variation, additive white noise and brightness variation. Moreover, we give a thorough discussion of the detection robustness regarding to video quality degradation. The robustness is composed by the detection accuracy on good-quality reference videos $A_\mathrm{ref}$ and the performance stability on distorted videos. Based on the rate of accuracy degradation and monotonicity criteria, we define the detection stability mathematically. We also propose an intuitive robustness presentation method named \emph{Robustness Quadrangle} which can be easily used to compare both the accuracies and stabilities between detectors. Usually, we treat the $A_\mathrm{ref}$ as the main attribute of robustness. However, when the $A_\mathrm{ref}$ of the compared detectors are close, as in many cases in practice, the stability measurement provides one more dimension to measure the robustness of the detectors.

With in-depth analysis of detection stability in Sec. \ref{sec:EvalResults}, we have the following findings: 1) To H.264 compression distortion, Lbp feature can be an important complement to HOG feature in pedestrian detection. 2) Detectors trained with low model height performs more stable when the spatial resolution of the video reduces. However, training with unreasonable low model height may result in detection accuracy decrease, and cannot work well with high resolution cases. Obviously, more careful studies are called for generalization of the learnt models. 3) LatSVM shows more promising stability \cite{felzenszwalb2010object} to additive gaussian noise compared with other evaluated detectors. And this is possibly resulted by the PCA-HOG, which only consider the main structures of HOG features after PCA.  4) HOG shows the best stability with brightness variation among all the features used by the evaluated detectors.

Based on the robustness evaluation, the following two crucial cases are important to improve the detection robustness in the future studies, distorted videos with quality under ``critical quality point'' and over-exposed videos with high brightness. To distorted videos with compression distortion, resolution reduction and additive white noise, the detection accuracy of most detectors does not gradually decrease when video quality drops, but decreases dramatically when the video quality reaches a critical point. Hence, efforts on extending the ``critical quality point'' could be one way to improve the detection stability. Compared with the pedestrian detection on low-brightness videos, detection on overexposed videos is an more challenging task and has been rarely studied.

As mentioned previously, much progress has been made to improve the detection accuracy in good-quality videos during the past decades. Comparing the evaluated pedestrian detectors in this study, the average miss rate on the high quality reference videos of \emph{DSurVD} has been reduced from around $90\%$ to $50\%$. However, quite less attention has been put on the research of the detection stability on surveillance videos with low quality. The in-depth analysis in this study have shown that there is still much room for improvement regarding to pedestrian detection in surveillance videos with low quality (often occurring in real-world situations).

\bibliographystyle{IEEEtranS}
\bibliography{refs}

\end{document}